\RequirePackage{snapshot}
\documentclass[twocolumn,oneside,a4paper,11pt]{article}

\newif\ifdebug %
\debugfalse %

\newif\ifblind %
\blindfalse
\newcommand{\authorsData}{{\large Hassan Hafez-Kolahi\\}
{\small Department of Computer Engineering, Sharif University of Technology, Tehran, Iran\\}
hafez@ce.sharif.edu\\
{\large Shohreh Kasaei*\\}
{\small Department of Computer Engineering, Sharif University of Technology, Tehran, Iran\\
skasaei@sharif.edu}}

\newcommand \authorsBio {\vskip 1.5cm
\textbf{Hassan Hafez-Kolahi}
was born in Mashhad, Iran, in 1989. He received the B.Sc. degree in Computer Engineering from Ferdowsi University of Mashhad in 2011, the M.Sc. degree in 2013 from Sharif University of Technology. He is currently a Ph.D. candidate at Sharif University of Technology. His research interests are in machine learning and information theory.

\vskip 0.5cm
\textbf{Shohreh Kasaei}
received the B.Sc. degree from the Department of Electronics, Faculty of Electrical and Computer Engineering, Isfahan University of Technology, Iran, in 1986, the M.Sc. degree from the Graduate School of Engineering, Department of Electrical and Electronic Engineering, University of the Ryukyus, Japan, in 1994, and the Ph.D. degree from Signal Processing Research Centre, School of Electrical and Electronic Systems Engineering, Queensland University of Technology, Australia, in 1998. She joined Sharif University of Technology since 1999, where she is currently a full professor and the director of Image Processing Laboratory (IPL). Her research interests are in 3D computer vision and image/video processing including dynamic 3D object tracking, dynamic 3D human activity recognition, virtual reality, 3D semantic segmentation, multi-resolution texture analysis, scalable video coding, image retrieval, video indexing, face recognition, hyperspectral change detection, video restoration, and fingerprint authentication.
}

\newcommand \AcknowledgmentText{We wish to thank Dr. Mahdieh Soleymani for her beneficial discussions and comments.}

\renewcommand{\footnote}[1]{} %
\usepackage[top=4cm, bottom=2cm, left=2cm, right=1.5cm]{geometry} %
\usepackage[framemethod=default]{mdframed}

\usepackage[pdftex,dvipsnames]{xcolor}  %
\usepackage{myforloop}
\usepackage{hhk_based_on_jmlr2e}%
\usepackage{amsmath,amssymb}
\usepackage[utf8]{inputenc}
\usepackage{graphicx}%
\usepackage{braket}
\usepackage{tikz}
\usetikzlibrary{automata,arrows,arrows.meta,positioning,calc,shapes,trees}

\usepackage{bbm}%
\usepackage{subcaption}
\usepackage{relsize}
\usepackage{float}
\floatstyle{plaintop}%
\restylefloat{table}
\usepackage{pgffor}

\usepackage[space]{grffile} %

\usepackage{datatool}

\usepackage{csvsimple}
\usepackage{xspace}
\usepackage{etoolbox} %
\usepackage{mathtools} %

\usepackage{xargs}                      %
\usepackage{hyperref}
\hypersetup{
    colorlinks,
    citecolor=black,
    filecolor=black,
    linkcolor=black,
    urlcolor=black
}

\newtheorem{mathdef}{Definition}

\usepackage{xhfill}

\usepackage[breakable, theorems, skins]{tcolorbox}

\usepackage{comment}

\specialcomment{commentInline}{\begin{tcolorbox}[breakable,left=0pt,right=0pt,top=0pt,bottom=0pt,colback=gray!20,colframe=gray!20,width=\dimexpr\textwidth\relax,enlarge left by=0mm,boxsep=5pt,arc=0pt,outer arc=0pt,] \noindent \xrfill[0.7ex]{1pt} PRIVATE COMMENT  \xrfill[0.7ex]{1pt}\\ 
 }{\par \noindent \xrfill[0.7ex]{1pt} END OF COMMENT  \xrfill[0.7ex]{1pt}\\ \end{tcolorbox}}

\ifdebug
\else
\excludecomment{commentInline}
\fi

\ifdebug
	\usepackage[colorinlistoftodos,prependcaption,textsize=tiny]{todonotes}
\else
	\usepackage[colorinlistoftodos,prependcaption,textsize=tiny,disable]{todonotes}
\fi

\newcommandx{\todoFirst}[2][1=]{\todo[linecolor=orange,backgroundcolor=orange!100,bordercolor=orange,#1]{TODO1: #2}}
\newcommandx{\todoSecond}[2][1=]{\todo[linecolor=orange,backgroundcolor=orange!50,bordercolor=orange,#1]{TODO2: #2}}
\newcommandx{\todoThird}[2][1=]{\todo[linecolor=orange,backgroundcolor=orange!25,bordercolor=orange,#1]{TODO3: #2}}
\newcommandx{\todoOptional}[2][1=]{\todo[linecolor=green,backgroundcolor=green!25,bordercolor=green,#1]{TODO3: #2}}

\newcommandx{\comUnsure}[2][1=]{\todo[linecolor=red,backgroundcolor=red!25,bordercolor=red,#1]{#2}}
\newcommandx{\comInfo}[2][1=]{\todo[linecolor=OliveGreen,backgroundcolor=OliveGreen!25,bordercolor=OliveGreen,#1]{#2}}

\ifdebug
	\usepackage[subject={Top1},author={\AA{}nsgar Lund},version=1]{pdfcomment}
	\defineavatar{HHK}{author=HHK,color=green,open=true}
	\newcommand{\commentTooltip}[1]{\pdftooltip[avatar=HHK]{\textcolor{red}{[hover to see comment]}}{#1 
	\textLF\textCR \textLF\textCR
	[This comment won't be present in the final version (use \textbackslash debugfalse).]}}
\else
	\newcommand{\commentTooltip}[1]{}
\fi

\newcommand{\dataset}{{\cal D}}

\makeatletter
\newcommand{\distas}[1]{\mathbin{\overset{#1}{\kern\z@\sim}}}%
\newsavebox{\mybox}\newsavebox{\mysim}
\newcommand{\distras}[1]{%
  \savebox{\mybox}{\hbox{\kern3pt$\scriptstyle#1$\kern3pt}}%
  \savebox{\mysim}{\hbox{$\sim$}}%
  \mathbin{\overset{#1}{\kern\z@\resizebox{\wd\mybox}{\ht\mysim}{$\sim$}}}%
}
\makeatother

\DeclareMathOperator{\EX}{\mathbb{E}}%
\DeclareMathOperator{\Indicator}{\mathbbm{1}}%

\newcommand\independent{\protect\mathpalette{\protect\independenT}{\perp}}
\def\independenT#1#2{\mathrel{\rlap{$#1#2$}\mkern2mu{#1#2}}}

\DeclarePairedDelimiterX{\infdivx}[2]{(}{)}{%
  #1\;\delimsize\|\;#2%
}

\newcommand{\KLdiv}{\text{KL}\infdivx}

\firstpageno{1}

\begin{document}

\twocolumn[
  \begin{@twocolumnfalse}

 \vspace*{-2cm}
Journal of Information Systems and Telecommunication

\vspace*{1.8cm}

\begin{center}

{\Huge \textbf{Information Bottleneck \\
and its Applications in Deep Learning\\}}

\vskip30pt

\authorsData

\end{center}

\section*{Abstract}
Information Theory (IT) has been used in Machine Learning (ML) from early days of this field. In the last decade, advances in Deep Neural Networks (DNNs) have led to surprising improvements in many applications of ML. The result has been a paradigm shift in the community toward revisiting previous ideas and applications in this new framework. Ideas from IT are no exception. One of the ideas which is being revisited by many researchers in this new era, is Information Bottleneck (IB); a formulation of information extraction based on IT. The IB is promising in both analyzing and improving DNNs. The goal of this survey is to review the IB concept and demonstrate its applications in deep learning. The information theoretic nature of IB, makes it also a good candidate in showing the more general concept of how IT can be used in ML. Two important concepts are highlighted in this narrative on the subject, i) the concise and universal view that IT provides on seemingly unrelated methods of ML, demonstrated by explaining how IB relates to minimal sufficient statistics, stochastic gradient descent, and variational auto-encoders, and ii) the common technical mistakes and problems caused by applying ideas from IT, which is discussed by a careful study of some recent methods suffering from them.

\vskip 0pt
\noindent \textbf{Keywords:}  Machine Learning; Information Theory; Information Bottleneck; Deep Learning; Variational Auto-Encoder.

\vskip 20pt

\end{@twocolumnfalse}
]

\section{Introduction}
The area of information theory was born by Shannon's landmark paper in 1948 \citep{Shannonmathematicaltheorycommunication1948}. One of the main topics of IT is communication; which is sending the \emph{information} of a source in such a way that the  receiver can decipher it. Shannon's work established the basis for quantifying the \emph{bits of information} and answering the basic questions faced in that communication.  On the other hand, one can describe the machine learning as the science of deciphering (decoding) the parameters of a true model (source), by considering a random sample that is generated by that model. In this view, it is easy to see why these two fields usually cross path each other. This dates back to early attempts of statisticians to learn parameters from a set of observed samples; which was later found to have interesting IT counterparts \citep{Kullbackinformationsufficiency1951}. Up until now, IT is used to analyze statistical properties of learning algorithms \citep{BassilyLearnersthatUse2018,VeraRoleInformationComplexity2018,NachumDirectSumResult2018}.

After the revolution of deep neural networks \citep{SrivastavaDropoutsimpleway2014}, the lack of theory that is able to explain its success \citep{ZhangUnderstandingdeeplearning2017} has motivated researchers to analyze (and improve) DNNs by using IT observations. The idea was first proposed by \cite{TishbyDeepLearningInformation2015} who made some connections between the information bottleneck method \citep{Tishbyinformationbottleneckmethod1999} and DNNs. Further experiments showed evidences that support the applicability of IB in DNNs \citep{Shwartz-ZivOpeningBlackBox2017}. After that, many researchers tried to use those techniques to analyze DNNs \citep{KhadiviFlowinformationfeedforward2016, AchilleEmergenceInvarianceDisentangling2017, Shwartz-ZivOpeningBlackBox2017, SaxeInformationBottleneckTheory2018} and subsequently improve them \citep{AlemiDeepVariationalInformation2017, KolchinskyNonlinearInformationBottleneck2017, AchilleInformationDropoutLearning2018}.

In this survey, in order to follow current research headlines, the main needed concepts and methods to get more familiar with the IB and DNN are covered. In Section \ref{sec:theEvolutionOfInformationExtraction}, the historical evolution of information extraction methods from classical statistical approaches to IB are discussed. Section \ref{sec:IBandDNN}, is devoted to the connections between IB and recent DNNs. In Section \ref{sec:beyondInformationBottleneck} another information theoretic approach for analyzing DNNs is introduced as an alternative to IB. Finally, Section \ref{sec:conclusion} concludes the survey.

\section{Evolution of Information Extraction Methods}
\label{sec:theEvolutionOfInformationExtraction}
A shared concept in statistics, information theory, and machine learning is defining and extracting the \emph{relevant information} about a target variable from observations. This general idea, was presented from the early days of modern statistics. It then evolved ever since taking a new form in each discipline which arose through time. As is expected from such a multidisciplinary concept, a complete understanding of it requires a persistent pursuit of the concept in all relevant fields. This is the main objective of this section. 
In order to make a clear view, the methods are organized in a chronological order with the emphasis on their \emph{cause} and \emph{effect}; i.e., \emph{why} each concept has been developed and \emph{what} has it added to the big picture. 

In the reminder of this section, first the notations are defined and after that the evolution of methods from sufficient statistics to IB is explained.

\subsection{Preliminaries and Notations}
\label{sec:noations}
Consider $X\in \mathcal{X}$ and $Y\in \mathcal{Y}$ as random variables with the joint distribution function of $p(x,y)$, where $\mathcal{X}$ and $\mathcal{Y}$ are called input and output spaces, respectively. Here, the realization of each Random Variable (r.v.) is represented by the same symbol in the lower case. The conditional entropy of $X$, given $Y$, is defined as 
$H(X | Y)= \EX [-\log p(X,Y)]$ and their Mutual Information (MI) is given by $I(X;Y)=\EX [\log \frac{p(X,Y)}{p(X)p(Y)}]$. There are also more technical definitions for MI allowing it to be used in cases that the distribution function $p(x,y)$ is singular\footnote{Note that this happens more often in neural networks. The reason is that in a simple setup of a layer with no noise, the output is computed deterministically from the input. Therefore, the joint distribution function of the input and output is singular; i.e., its support is a low dimensional manifold in which the deterministic relation is satisfied.} \citep{KolmogorovShannontheoryinformation1956, CoverKolmogorovContributionsInformation1989}. An important property of MI is that it is invariant under bijective transforms $f$ and $g$; i.e., $I(X;Y)=I(f(X),g(Y))$ \citep{CoverElementsinformationtheory2012}.

A noisy channel is described by a conditional distribution function $p(\tilde{x}|x)$, in which $\tilde{x}\in \mathcal{\tilde{X}}$ is the noisy version of $X$. In the rate distortion function, the distortion function $d: \mathcal{X}\times\widetilde{\mathcal{X}}\to \mathbb{R}$ is given and the minimum required bit-rate for a fixed expected distortion is studied. Then
\begin{equation}
R(D)=\min_{\substack{p(\tilde{x}|x)\\ s.t. \EX [d(X,\tilde{X})]\le D}} I(\tilde{X};X).
\end{equation}

\subsection{Minimal Sufficient Statistics}
\label{sec:MSS}
A core concept in statistics is defining the \emph{relevant} information about a target $Y$ from observations $X$. One of the first mathematical formulations proposed for measuring the relevance, is the concept of \emph{sufficient statistic}. This concept is defined below \citep{RAFishermathematicalfoundationstheoretical1922}. 
\begin{mathdef}
(Sufficient Statistics).
Let $Y\in \mathcal{Y}$ be an unknown parameter and $X\in \mathcal{X}$ be a random variable with conditional probability distribution function $p(x|y)$. Given a function $f:\mathcal{X} \to \mathcal{S}$, the random variable $S=f(X)$ 
is called a sufficient statistic for $Y$ if 
\begin{equation}
\begin{split}
\forall & x \in \mathcal{X},y \in \mathcal{Y}: \\ &P(X=x|Y=y,S=s)=P(X=x|S=s).
\end{split}
\end{equation}

\end{mathdef}
In other words, a sufficient statistic captures all the information about $Y$ which is available in $X$. This property is stated in the following theorem \citep{ShamirLearninggeneralizationinformation2010, Kullbackinformationsufficiency1951}.

\begin{theorem}
\label{theorem:sufficientStatisticInformationForm}
Let $S$ be a probabilistic function of $X$. Then, $S$ is a sufficient statistic for $Y$ if and only if (iff) 
\begin{equation}
I(S;Y)=I(X;Y).
\end{equation}
\end{theorem}

Note that in many classical cases that one encounters in point estimation, it is assumed that there is a family of distribution functions that is parameterized by an unknown parameter $\theta$ and furthermore $N$ Independent and Identically Distributed (i.i.d.) samples of the target distribution function are observed. This case fits the definition by setting $Y=\theta$ and considering the high dimensional random variable $X=\{X^{(i)}\}_{i=1}^N$ that contains all observations.

A simple investigation shows that the sufficiency definition includes the trivial identity statistic $S=X$. Obviously, such statistics are not helpful, as copying the whole signal cannot be called "extraction" of relevant information. Consequently, one needs a way to restrict the sufficient statistic from being wasteful in using observations. To address this issue, authors of  \cite{LehmannCompletenessSimilarRegions1948} introduced the notion of \emph{minimal} sufficient statistics. This concept is defined below.
\begin{mathdef}
(Minimal Sufficient Statistic) A sufficient statistic $S$ is said to be minimal if it is a function of all other sufficient statistics 
\begin{equation}
\forall T; T \text{is sufficient statistic} \Rightarrow \exists g; S=g(T).
\end{equation}
\end{mathdef}
It means that a Minimal Sufficient Statistic (MSS) has the coarsest partitioning of the input variable $X$. In other words, an MSS tries to group the values of $\mathcal{X}$ together in as few number of partitions as possible, while making sure that there is no \emph{information}
loss in the process.

The following theorem describes the relation between minimal sufficient statistics and mutual information\citep{ShamirLearninggeneralizationinformation2010}\todoThird{I found the proof of the theorem just in Tishby's work, I have doubts that there are subtle mathematical difficulties not captured by them. (Specially the case where $S$ is stochastic, but the way I presented the theorem, I guess does not have that problem).}.
\begin{theorem}
\label{theorem:minimalSufficientStatisticInformationForm}
Let $X$ be a sample drawn from a distribution function that is determined by the random variable $Y$. The statistic $S$ is an MSS for $Y$ iff it is a solution of the optimization process
\begin{equation}
\label{eq:MSS_IT_Formulation}
\min_{T: \text{sufficiet statistic}} I(X;T).
\end{equation}

\end{theorem}
By using Theorem \ref{theorem:sufficientStatisticInformationForm}, the constraint of this optimization problem can be written by information theory terms, as
\begin{equation}
\label{eq:MSS_IT_FormulationFinal}
\min_{T: I(T;Y)=I(X;Y)} I(X;T).
\end{equation}

It shows that MSS is the statistic that have all the available information about $Y$, while retaining the minimum possible information about $X$. In other words, it is the best compression of $X$, with zero information loss about $Y$.

\begin{table*} %
\centering
\resizebox{0.72\linewidth}{!}{
\begin{minipage}{\textwidth}%
\begin{tabular}{|c|c|c|}
\cline{2-3} 
\multicolumn{1}{c|}{}

& \rule{0pt}{4ex} \textbf{ Markov Chain }
& \textbf{ Data Processing Inequality }

\\[5pt]
\hline 
\rule{0pt}{7ex}    
\raisebox{1.3\height}{ \textbf{Statistic}} & 
\resizebox{0.4\linewidth}{!}{
	\begin{tikzpicture}[->, >=stealth', auto, semithick, node distance=3cm]
	\tikzstyle{every state}=[fill=white,draw=black,thick,text=black,scale=1]
	\node[state]    (A)   {$Y$};
	\node[state,fill=gray!40]    (B)[right of=A]   {$X$};
	\node[state]    (C)  [right of=B]                   {S};
	
	\path
	(A) edge[-]   node{}     (B)
	(B) edge[-]   node{}  	 (C);
	\end{tikzpicture}
}
&

\raisebox{1.3\height}{ $I(S;Y) \le I(X;Y)$}

\\[5pt]
\hline 
\rule{0pt}{7ex}    
\raisebox{1.3\height}{ \textbf{Sufficient}} & 
\resizebox{0.4\linewidth}{!}{
	\begin{tikzpicture}[->, >=stealth', auto, semithick, node distance=3cm]
	\tikzstyle{every state}=[fill=white,draw=black,thick,text=black,scale=1]
	\node[state]    (A)   {$Y$};
	\node[state]    (B)[right of=A]   {$SS$};
	\node[state,fill=gray!40]    (C)  [right of=B]                   {$X$};
	
	\path
	(A) edge[-]   node{}     (B)
	(B) edge[-]   node{}  	 (C);
	\end{tikzpicture}
}
&

\raisebox{1.3\height}{ $I(SS;Y) \ge I(X;Y)$}

\\[5pt]
\hline 
\rule{0pt}{7ex}    
\raisebox{1.3\height}{ \textbf{Minimal}} & 
\resizebox{0.4\linewidth}{!}{
	\begin{tikzpicture}[->, >=stealth', auto, semithick, node distance=3cm]
	\tikzstyle{every state}=[fill=white,draw=black,thick,text=black,scale=1]
	\node[state,fill=gray!40]    (A)   {$X$};
	\node[state]    (B)[right of=A]   {$SS$};
	\node[state]    (C)  [right of=B]                   {$MSS$};
	
	\path
	(A) edge[-]   node{}     (B)
	(B) edge[-]   node{}  	 (C);
	\end{tikzpicture}
}
&

\raisebox{1.3\height}{ $\forall \; SS: I(MSS;X) \le I(SS;X)$}

\\ 
\hline 
\end{tabular} 

\caption{Markov chains corresponding to conditions that form a Minimal Sufficient Statistic, along with its enforced information inequality.}
\label{table:MSSMarkovChainsAndInequalities}
\end{minipage}}
\end{table*}

In Table \ref{table:MSSMarkovChainsAndInequalities}, the components of MSS are presented in a concise way by using Markov chains. Note that these Markov chains should hold for every possible statistic $S$, sufficient statistic $SS$, and minimal sufficient tatistic $MSS$. By these three Markov chains and the information inequalities corresponding to each, it is easy to verify Theorems \ref{theorem:sufficientStatisticInformationForm} and \ref{theorem:minimalSufficientStatisticInformationForm}. By using the two first inequalities, is easily proved that $I(SS;Y)=I(SS;X)$ . The last inequality shows that $MSS$ should be the $SS$ with minimal $I(SS;X)$.

In most practical problems where $X=\{(X^{(i)}\}_{i=1}^N$ is an $N$-dimensional data, one hopes to find a (minimal) sufficient statistic $S$ in such a way that its dimension does not depend on $N$. %
Unfortunately, it is found to be impossible for almost all distributions (except the ones belonging  to the exponential family) \citep{ShamirLearninggeneralizationinformation2010,Koopmandistributionsadmittingsufficient1936}.

\subsection{Information Bottleneck}
\label{sec:IB}

To tackle this problem, Tishby presented the IB method to solve the Lagrange relaxation of the optimization function  
\eqref{eq:MSS_IT_FormulationFinal}, by\citep{Tishbyinformationbottleneckmethod1999}
\begin{equation}
\label{eq:IB}
\min_{p(\tilde{x}|x)}I(\tilde{X};X)-\beta I(\tilde{X};Y)
\end{equation}
where $\tilde{X}$ is the representation of $X$, and $\beta$ is a positive parameter that controls the trade-off between the compression and preserved information about $Y$. For $\beta<=1$, the trivial case where $\tilde{X} \independent X$ is a solution. The reason is that the data processing inequality enforces $I(\tilde{X};X)\ge I(\tilde{X};Y) = ((1-\beta)+\beta) I(\tilde{X};Y)$. Therefore, the value $(1-\beta)I(\tilde{X};Y)$ is a lower bound for the objective function of optimization problem \eqref{eq:IB}. For $\beta\ge 1$, this lower bound is minimized by setting $I(\tilde{X};Y)=0$. It is achieved by simply choosing $I(\tilde{X};X)=0$. 

As such, the solution starts from $I(\tilde{X};X)=I(\tilde{X};Y)=0$, and by increasing $\beta$, both $I(\tilde{X};X)$ and $I(\tilde{X};Y)$ are increased. At the limit, $\beta \to \infty$, this optimization function is equivalent to \eqref{eq:MSS_IT_Formulation} \citep{ShamirLearninggeneralizationinformation2010}. Note that in IB, the optimization function is performed on conditional distribution functions $p(\tilde{x}|x)$. Therefore, the solution is no longer restricted to deterministic statistics $T=f(X)$. In general, the optimization function \eqref{eq:IB} does not necessarily have a deterministic solution. This is true even for simple cases with two binary variables \citep{AchilleInformationDropoutLearning2018}. 
The IB provides a quite general framework with many extensions
\commentTooltip{It can be used for both discrete and continuous variables \citep{ChechikInformationbottleneckGaussian2005}}
(there are variations of this method for more than one variable \citep{FriedmanMultivariateinformationbottleneck2001}). But, since there is no evident connection between these variations and DNNs, they are not covered in this survey.

\commentTooltip{The original application of IB was the information theoretic clustering; which means to find the (soft) partitions of $X$ values that are informative. This idea was later extended to cases with more than two variables\citep{FriedmanMultivariateinformationbottleneck2001}.}

Tishby et al. showed that IB has a nice rate-distortion interpretation, using the distortion function $d(x,\tilde{x})=\KLdiv{p(y|x)}{p(y|\tilde{x})}$ \citep{Gilad-Bachrachinformationtheoretictradeoff2003}. It should be noted that this does not exactly conform to the classical rate-distortion settings, since here the distortion function implicitly depends on the $p(\tilde{x}|x)$ which is being optimized. They provided an algorithm similar to the well-known Blahut-Arimoto rate-distortion algorithm \citep{BlahutComputationchannelcapacity1972, Arimotoalgorithmcomputingcapacity1972} to solve the IB problem. %

Till now, it was considered that the joint distribution function of $X$ and $Y$ is known. But, it is not the case in ML. In fact, if one knows the joint distribution function, then the problem is usually as easy as computing an expectation on the conditional distribution function; e.g.,  $f(x)=\EX_{p(y|x)}[Y]$ for regression and $E_{p(y|x)}[\Indicator(Y=c)]; c \in \mathcal{Y}$ for classification. Arguably, one of the main challenges of ML is to solve the problem when one has the access to the distribution function through a finite set of samples.

Interestingly, it was found that the value of $\beta$, introduced as a Lagrange relaxation parameter in \eqref{eq:IB}, can be used to control the bias-variance trade-off in cases for which the distribution function is not known and the mutual information is just estimated from a finite number of samples. It means that instead of trying to reach the MSS by setting $\beta \to \infty$, when the distribution function is unknown, one should settle for a $\beta^* < \infty$ which gives the best bias-variance trade-off \cite{ShamirLearninggeneralizationinformation2010}. \label{sec_part:discussionAboutBiasVarianceAndBestBeta}
The reason is that the error of estimating the mutual information from finite samples is bounded by $O(\frac{|\widetilde{\mathcal{X}}|\log m}{\sqrt{m}})$, where $|\widetilde{\mathcal{X}}|$ is the number of possible values that the random variable $\tilde{X}$ can take (see Theorem 1 of \cite{ShamirLearninggeneralizationinformation2010}).  The $|\widetilde{\mathcal{X}}|$ has a direct relation with $\beta$: small $\beta$ means more compressed $\tilde{X}$, meaning that less distinct values are required to represent $\tilde{X}$. 
This is in line with the general rule that simpler models generalize better. As such, there are two opposite forces in play, one trying to increase $\beta$ to make the Lagrange relaxation of optimization function \eqref{eq:IB} to be more accurate, while the other tries to decrease $\beta$ in order to control the finite sample estimation errors of $I(\tilde{X};X)$ and $I(\tilde{X};Y)$.
The authors of \cite{ShamirLearninggeneralizationinformation2010} also tried to make some connections between the IB and the classification problem. Their main argument is that in equation \eqref{eq:IB}, $I(\tilde{X}|Y)$ can be considered as a proxy for the classification error. They showed that if two conditions are met, the miss-classification error is bounded from above by $I(\tilde{X}|Y)$ . These conditions are: i) the classes have equal probability, and ii) each sample is composed of a lot of components (as in the document (text) classification setting). The latter is equivalent to the general technique in IT where one can neglect small probabilities when dealing with \emph{typical} sets. They also argued that $I(\tilde{X};X)$ is a regularization term that controls the generalization-complexity trade-off.

The main limitation of their work is that they considered both $X$ and $Y$ to be discrete. This assumption is violated in many applications of ML; including image and speech processing. While there are extensions to IB allowing to work with continuous random variables \citep{ChechikInformationbottleneckGaussian2005}, their finite sample analysis and the connections to ML applications are less studied.

\section{Information Bottleneck and Deep Learning}
\label{sec:IBandDNN}
After the revolution of DNNs, which started by the work of 
\cite{KrizhevskyImagenetclassificationdeep2012}, in various areas of ML the state-of-the-art algorithms were beaten by DNN alternatives.
While most of the ideas used in DNNs existed for decades, the recent success attracted unprecedented attention of the community. In this new paradigm, both practitioners and 
theoreticians found new ideas to either use DNNs to solve specific problems or use previous theoretical tools to understand DNNs. 

Similarly, the interaction of IB and DNN in the literature can be divided in two main categories. The first is to use the IB theories in order to analyze DNNs and the other is to use the ideas from IB to improve the DNN-based learning algorithms. The remaining of this section is divided based on these categories. 

Section \ref{sec:IBandSGD} is devoted to the application of IB in analyzing the usual DNNs, which is mainly due to the conjecture that Stochastic Gradient Descent, the de facto learning algorithm used for DNNs, implicitly solves an IB problem. In Section \ref{sec:IBandVAE}, the practical applications of IB for improving DNNs and developing new structures are discussed. The practical application is currently mostly limited to Variational Auto-Encoders (VAEs).

\subsection{Information Bottleneck and Stochastic Gradient Descent}
\label{sec:IBandSGD}
From theoretical standpoint, the success of DNNs is not completely understood. The reason is that many learning theory tools analyze models with a limited \emph{capacity} and find inequalities restricting the deviation of train test statistics. But, it was shown that commonly used DNNs have huge capacities that make such theoretical results to be inapplicable \citep{ZhangUnderstandingdeeplearning2017, VeraRoleInformationComplexity2018}.\commentTooltip{In Vera's work it briefly stated a few ML techniques and said they are failed} In recent years, there were lots of efforts to mathematically explain the generalization capability of DNNs by using variety of tools. They range from attributing it to the way that the SGD method automatically finds flat local minima (which are \emph{stable} and thus can be well generalized) \citep{Keskarlargebatchtrainingdeep2016, HochreiterFlatminima1997, ChaudhariEntropysgdBiasinggradient2016, HardtTrainfastergeneralize2015}, to efforts trying to relate the success of DNNs to the special class of hierarchical functions that they generate \citep{PoggioWhyWhenCan2016}. 
Each of these categories has its critics and thus the problem is still under debate (e.g., \cite{DinhSharpminimacan2017}  argues that flatness can be changed arbitrarily by re-parametrization and the direct relation between generalization and flatness is not generally true). In this survey, the focus is on a special set of methods that try to analyze DNNs by information theory results (see \cite{VidalMathematicsDeepLearning2017} for a broader discussion).

\begin{figure*}
\centering
\begin{subfigure}{0.5\textwidth}

\raisebox{0.6cm}{
\includegraphics[width=0.9\linewidth]{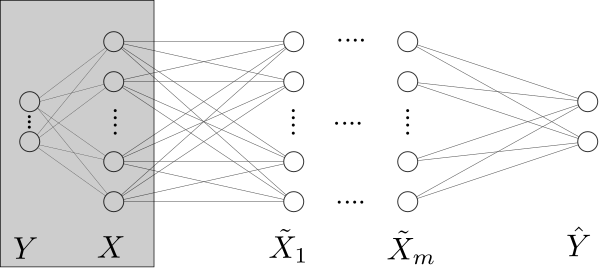}
}
\caption{}
\label{fig:NNMC}
\end{subfigure}%
\begin{subfigure}{0.5\textwidth}

\includegraphics[width=0.9\linewidth]{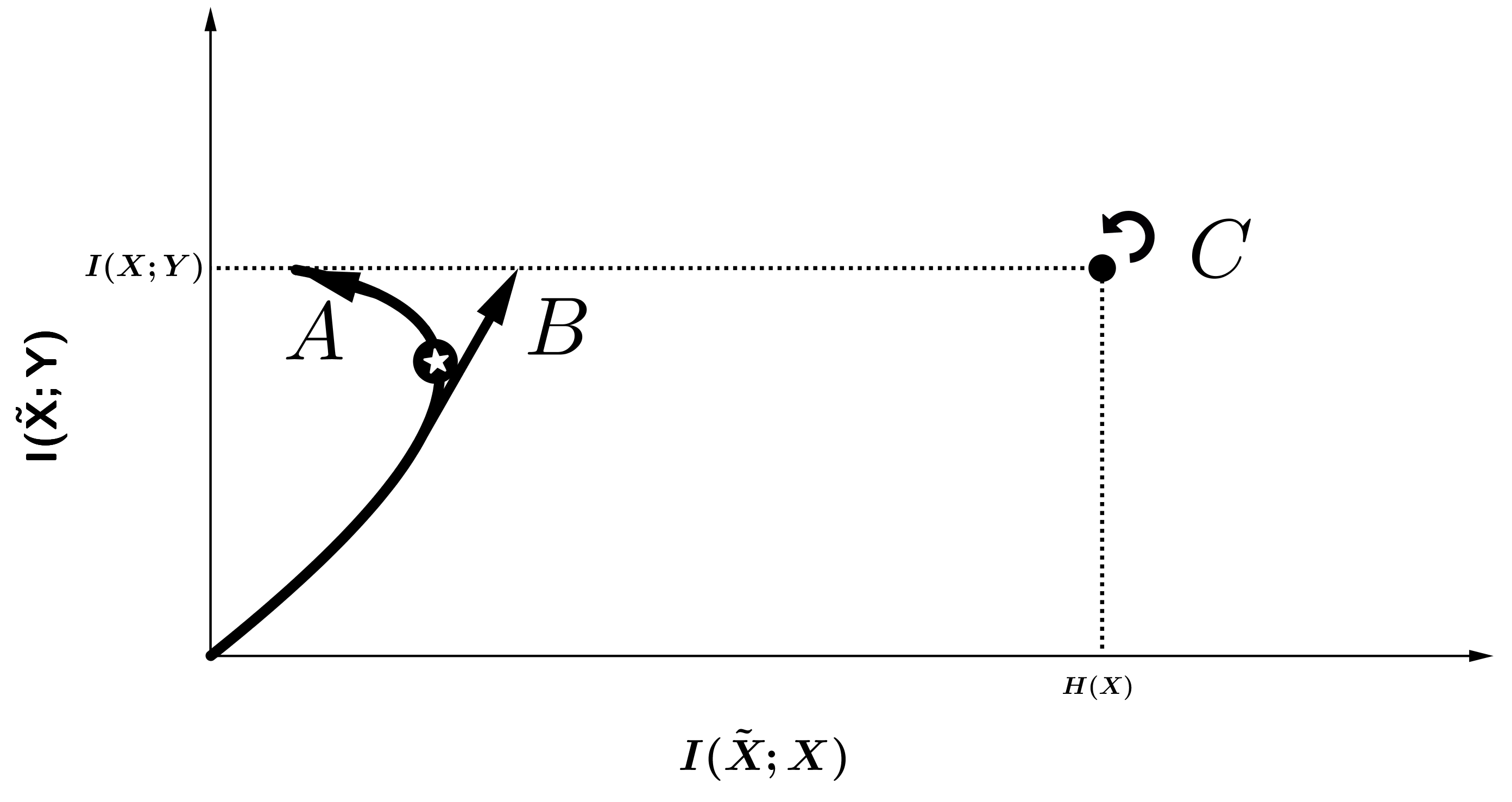}
\caption{}
\label{fig:informationPlaneDiagram}
\end{subfigure}
\caption{Information plane diagram of DNNs. (a) Markov chain representation of a DNN with $m$ hidden layers. [Note that the predicted label $\hat{Y}$ has access to $Y$ only through $X$.] (b) Path hidden layers undergo during SGD training in information plane. Three possible paths under debate by authors are represented by $A$, $B$, and $C$.}
\label{fig:NNMCandInformationPlaneDiagram}
\end{figure*}

Tishby et al. used ideas from IB to formulate the goal of deep learning as an information theoretic trade-off between compression and prediction \citep{TishbyDeepLearningInformation2015}. In that view, an NN forms a Markov chain of representations, each trying to refine (compress) the representation while preserving the information about the target. Therefore, they argued that DNN is automatically trying to solve an IB problem and the last layer is the optimal representation $\tilde{x}$ that is to be found. Then, they used the generalization theories of IB (discussed in \ref{sec:IB}) to explain the success of DNNs. One of their main contributions is the idea to use the \emph{information plane} diagrams showing the inside performance of a DNN (see Figure \ref{fig:informationPlaneDiagram}). The information plane is a 2D diagram with  $I(\tilde{X};X)$ and $I(\tilde{X};Y)$ as the $x$ and $y$ axis, respectively. In this diagram , each layer of the network is represented by a point that shows how much information it contains about the input and output.

Later, they also practically showed that in learning DNNs by a simple SGD (without regularization or batch normalization), the compression actually happens \citep{Shwartz-ZivOpeningBlackBox2017}. The Markov chain representation that they used and their results are shown in Figure \ref{fig:NNMCandInformationPlaneDiagram}.
As the SGD proceeds, by tracking each layer on the information plane, they reported observing the path $A$ in Figure \ref{fig:informationPlaneDiagram}. In this path, a deep hidden layer starts from point $(0,0)$. The justification is that at the beginning of SGD, where all weights are chosen randomly, the hidden layer is meaningless and does not hold any information about either of $X$ or $Y$. During the training phase, as the prediction loss is minimized,  $I(\tilde{X};Y)$ is expected to increase (since the network uses $\tilde{X}$ to predict the label, and its success depends on how much information $\tilde{X}$ has about $Y$).
But, changes in $I(\tilde{X};X)$ are not easy to predict. The surprising phenomena that they reported is that at first $I(\tilde{X};X)$ increases (called the learning phase). But, at some point a phase transition happens (presented by a star in Figure \ref{fig:informationPlaneDiagram}) and $I(\tilde{X};X)$ starts to decrease (called the compression phase). It is surprising because the minimized loss in deep learning does not have any \emph{compression} term. By experimental investigations, they also found that compression happens in later steps of SGD when the empirical error is almost zero and the gradient vector is dominated by its noisy part (i.e., observing a small gradient mean but a high gradient variance). By this observation, they argued that after reaching a low empirical error, the noisy gradient descent forms a diffusion process which approaches the stationary distribution that maximizes the entropy of the weights, under the empirical error constraint. They also explained how deeper structures can help SGD to faster approach to the equilibrium. In summary, their results suggested that the reason behind the DNN success, is that it automatically learns short descriptions of samples, which in turn controls the capacity of models. They reported their results for both synthesis datasets (true mutual information values) and real datasets (estimated mutual information values).

Saxe et al. \citep{SaxeInformationBottleneckTheory2018}  further investigated this phenomena on more datasets and different kinds of activation functions. They observed the compression phase just in cases for which a saturating activation function is used (e.g., $sigmoid$ or $tanh$). They argued that the explanation of diffusion process is not adequate to explain all different cases; e.g., for Relu activation which is commonly used in the literature, they usually could not see any compression phase (path $B$ in Figure \ref{fig:informationPlaneDiagram}).
It should be noted that their observations do not take the effect of compression completely out of picture, rather they just reject the universal existence of an explicit compression phase at the end of the training phase. As shown in Figure \ref{fig:informationPlaneDiagram}, even though there is no compression phase in Path B, the resulting representation is still compressed compared to $X$. This compression effect can be attributed to the initial randomness of the network rather than an explicit compression phase.
They also noticed that the way that the mutual information is estimated is crucial in the process. One of the usual methods for mutual information estimation is \emph{binning}. In that approach, the bin size is the parameter to be chosen. They showed that for small enough bin sizes, if the precision error of arithmetic calculations is not involved, there will not be any information loss to begin with (Path $C$ in Figure \ref{fig:informationPlaneDiagram}). The reason is that when one projects a finite set of distinct points to a random lower dimensional space, the chance that any two points get mixed is zero. Even though this problem is seemingly just an estimation error caused by a low number of samples in each bin (and thus does not invalidate synthesis data results of \cite{Shwartz-ZivOpeningBlackBox2017}), it is actually connected to a more fundamental problem. If one removes the binning process and deals with true values of mutual information, serious problems will arise when using IB to study common DNNs on continuous variables. The problem is that in usual DNNs, for which the hidden representation has a deterministic relation with inputs, the IB functional of optimization \eqref{eq:IB} is infinite for almost all weight matrices and thus the problem is ill-posed. This concept was further investigated in \cite{AmjadHowNotTrain2018}. 

Even though the problem was not explicitly addressed until recently, there are two approaches used by researchers that automatically tackle this problem. As mentioned before, the first approach, used by \cite{TishbyDeepLearningInformation2015}, applies binning techniques to estimate the mutual information. This is equivalent to add a (quantization) noise, making the IB functional limited. But, in this way, the noise is added just for the analysis process and does not affect the NN. As noted by \cite{SaxeInformationBottleneckTheory2018}, unfortunately some of the advertised characteristics of mutual information, namely the information inequality for layers and the invariance on reparameterization of the weights, does not hold any more.

The second approach is to explicitly add some noise to the layers and thus make the NN truly stochastic. This idea was first discussed by \cite{Shwartz-ZivOpeningBlackBox2017} as a way to make IB to be biased toward simpler models (as is usually desired in ML problems). It was later found that there is a direct relationship between the SGD and variational inference \citep{ChaudhariStochasticgradientdescent2017}. On the other hand, the variational inference has a "noisy computation" interpretation \citep{AchilleInformationDropoutLearning2018}. These results showed that the idea of using stochastic mappings in NNs has been used much earlier than the recent focus on IB interpretations. In the light of this connection, researchers tried to propose new learning algorithms based on IB in order to more explicitly take the compression into account. These ideas are strongly connected to Variational Auto-Encoders (VAEs) \cite{KingmaAutoencodingvariationalbayes2013}. The \emph{denoising} auto-encoders \cite{BengioGeneralizeddenoisingautoencoders2013,ImDenoisingCriterionVariational2017} also use an explicit noise addition and thus can be studied in the IB framework. The next section is devoted to the relation between IB and VAE which recently has been a core concept in the field . 

\subsection{Information Bottleneck and Variational Auto-Encoder}
\label{sec:IBandVAE}
Achille et al.  \cite{AchilleInformationDropoutLearning2018} introduced the idea of \emph{information dropout} in correspondence to the commonly used dropout technique \citep{SrivastavaDropoutsimpleway2014}. Starting from the loss functional in the optimization function \eqref{eq:IB} and noting that $I(\tilde{X};Y)=H(Y)-H(Y|\tilde{X})$, one can rewrite the problem as
\begin{equation}
\label{eq:IB_soatto_form}
\min_{p(\tilde{x}|x)}I(X;\tilde{X})+\beta H(Y|\tilde{X}).
\end{equation}
Moreover, the terms can be expanded as per sample loss of
\begin{eqnarray}
H(Y|\tilde{X})&=&
\EX_{p(x,y)}\left[\EX_{p(\tilde{x}|X)}[-\log(p(Y|\tilde{X}))]  \right]\nonumber \\
\label{eq:MI_KLform}
I(X;\tilde{X})&=& \EX_{p(x)} \left[ \KLdiv{p(\tilde{x}|X)}{p(\tilde{x})} \right]
\end{eqnarray}
where KL denotes the Kullback-Leibler divergence. The expectations in these two equations can be estimated by a sampling process. For distribution functions $p(x)$ and $p(x,y)$, the training samples $\dataset=\{(x^{(i)},y^{(i)})\}_{i=1}^N$ are already given. Therefore,  the loss function of IB can be approximated as
\begin{equation}
\begin{split}
\label{eq:IB_like_variational}
\mathcal{L}=\frac{1}{N}\sum_{i=1}^{N} \EX&_{p(\tilde{x}|x^{(i)})}[-\log(p(y^{(i)}|\tilde{x}))] \\
&+\beta \KLdiv{p(\tilde{x}|x^{(i)})}{p(\tilde{x})}.
\end{split}
\end{equation}

It is worth noting that if we let $\tilde{x}$ to be the output of NN, the first term is the cross entropy (which is the loss function usually used in deep learning). The second term acts like a regularization term to prevent the conditional distribution function $p(\tilde{x}|x)$ from being too dependent to the value of $x$. As noted by  \citep{AchilleInformationDropoutLearning2018}, this formulation reveals interesting resemblance to Variational Auto-Encoder (VAE) presented by \cite{KingmaAutoencodingvariationalbayes2013}. The VAE tries to solve the unsupervised problem of reconstruction, by modeling the process which has generated each data from a (simpler) random variable $\tilde{x}$ with a (usually fixed) prior $p_0(\tilde{x})$. The goal is to find the generative distribution function $p_{\theta(x|\tilde{x})}$ and also a variational approximation $p_\phi(\tilde{x}|x)$. This is done by minimizing the variational lower-bound of the marginal log-likelihood of the training data, given by \cite{AchilleInformationDropoutLearning2018}
\begin{equation}
\begin{split}
\label{eq:variational}
\mathcal{L}_{\theta,\phi}=
\frac{1}{N}\sum_{i=1}^{N} &\EX_{p_\phi(\tilde{x}|x^{(i)})}[-\log(p_\theta(x^{(i)}|\tilde{x}))]\\
&+\KLdiv{p_\phi(\tilde{x}|x^{(i)})}{p_0(\tilde{x})}.
\end{split}
\end{equation}
Comparing this with equation \eqref{eq:IB_like_variational}, it is evident that VAE can be considered as an estimation for a special case of IB when: i) $Y=X$, ii) $\beta=1$, iii) the prior distribution function is fixed $p(\tilde{x})=p_0(\tilde{x})$, and iv) the distribution functions $p(\tilde{x}|x)$ and $p(x|\tilde{x})$ are parameterized by $\phi$ and $\theta$, respectively. These parameters are optimized separately as suggested by the variational inference (note that in IB, the attention is on $p(\tilde{x}|x)$, and assuming that $p(x,y)$ is given, the values of $p(\tilde{x})$ and $p(y|\tilde{x})$ are determined from that). It is worth noting that the ii and iii restrictions are crucial. The reason is that just setting $X=Y$ and $\beta=1$, without any other restrictions, would make the objective function \eqref{eq:IB} to be a constant, making every $p(\tilde{x}|x)$ to be a solution. Even if $\beta \ne 1$, the trivial loss function $(1-\beta)I(\tilde{X};X)$ is obtained which is minimized either for $x=\tilde{x}$ (when $\beta>1$) or $x \independent \tilde{x}$ (when $\beta<1$). Neither of these solutions is desired in representation learning (for another view on this matter, see the discussion of \cite{AlemiFixingBrokenELBO2018} on "feasible" vs "realizable" solutions). 

A similar variational approach, is used to solve the IB optimization process \eqref{eq:IB_like_variational}, which is a more general setting with $\beta\ne 1$ and $X \ne Y$ \citep{AchilleInformationDropoutLearning2018}. 

\newmdenv[linecolor=gray!10,backgroundcolor=gray!10]{myframe}
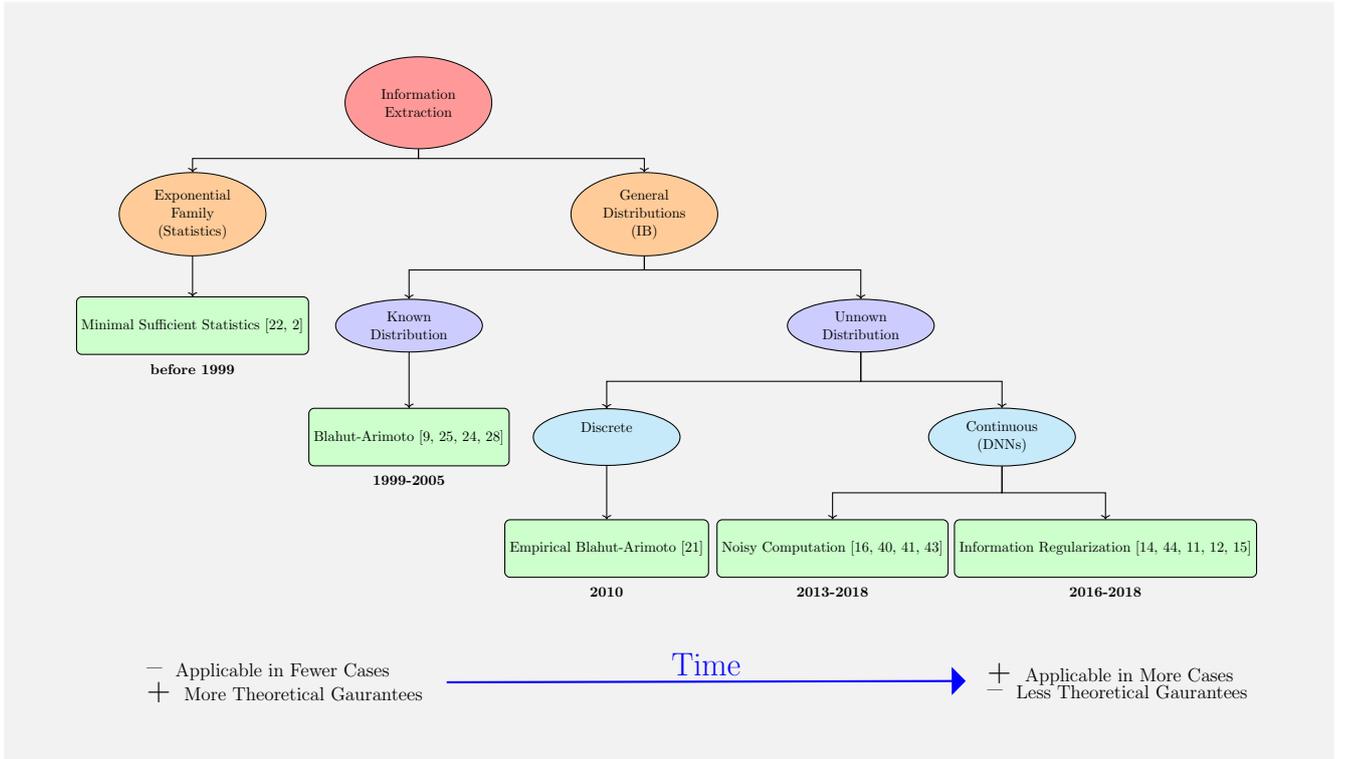
\begin{figure*}[t]
\begin{myframe}
\centering
\tikzstyle{block} = [rectangle, draw, rounded corners, minimum height=4em,fill=green!20]
\tikzstyle{line} = [draw,thick, -latex']
\tikzstyle{cloud} = [draw, ellipse, text width=2.5cm, text centered]
\tikzstyle{edge from parent}=[->,thick,draw]
\resizebox{1\linewidth}{!}{%

\setlength{\fboxsep}{30pt}%
\setlength{\fboxrule}{0pt}%
\fbox{
\begin{tikzpicture}[auto,edge from parent fork down]%
\tikzstyle{level 1}=[sibling distance=180mm,level distance=18ex] 
\tikzstyle{level 2}=[sibling distance=55mm,level distance=18ex] 

\node [cloud,fill=red!40,minimum height=70,minimum width=100] (cst) {Information Extraction}
	child{node [cloud,fill=orange!40,xshift=3cm] (pmt) {Exponential\\Family (Statistics)}
	    child{node [block,fill=green!20] (opm) {Minimal Sufficient Statistics \cite{LehmannCompletenessSimilarRegions1948,Kullbackinformationsufficiency1951}}}
	}
	child{node [cloud,fill=orange!40,xshift=-3cm] (gd) {General\\ Distributions (IB)}
	    child{node [cloud,fill=blue!20,xshift=-3.5cm] (kd) {Known Distribution}
			child{node [block] (ibba) {Blahut-Arimoto \cite{Tishbyinformationbottleneckmethod1999,Gilad-Bachrachinformationtheoretictradeoff2003,FriedmanMultivariateinformationbottleneck2001,ChechikInformationbottleneckGaussian2005}}}
		}
	    child{node [cloud,fill=blue!20,xshift=3cm] (gp) {Unnown Distribution}
			child{node [cloud,fill=cyan!20,xshift=-4cm] (dd) {Discrete\\\vphantom{Empty}}
				child{node [block] (eba) {Empirical Blahut-Arimoto \cite{ShamirLearninggeneralizationinformation2010}}}
			}
			child{node [cloud,fill=cyan!20,xshift=1cm] {Continuous (DNNs)}
				child{node [block,xshift=-1.75cm] (nc){
					Noisy Computation \cite{AchilleInformationDropoutLearning2018,BengioGeneralizeddenoisingautoencoders2013,ImDenoisingCriterionVariational2017,ChenVariationallossyautoencoder2016}}
				}
				child{node [block](ir){
					Information Regularization \cite{AlemiDeepVariationalInformation2017,HuangFlowRenyiinformation2016,KhadiviFlowinformationfeedforward2016,AchilleEmergenceInvarianceDisentangling2017,KolchinskyNonlinearInformationBottleneck2017}}
				}
			}
		}
	};
\node (time1) [below=of opm, yshift=25]{\textbf{before 1999}};
\node (time2) [below=of ibba, yshift=25]{\textbf{1999-2005}};
\node (time3) [below=of eba, yshift=25]{\textbf{2010}};
\node (time4) [below=of nc, yshift=25]{\textbf{2013-2018}};
\node (time5) [below=of ir, yshift=25]{\textbf{2016-2018}};
\node (AuxBelowTimesLeftmost) [below=of time5, yshift=-0.5cm, xshift=-1.3\textwidth,align=left]{{\Huge --} \; \Large Applicable in Fewer Cases\\ {\Huge +} \; \Large More Theoretical Gaurantees};
\node (AuxBelowTimesRightmost) [below=of time5, yshift=-0.5cm,xshift=0.1cm,align=left]{ {\Huge \;\;+} \; \Large Applicable in More Cases\\ {\Huge \;\;--} \; \Large Less Theoretical Gaurantees};
\draw[blue,ultra thick, arrows={-Triangle[angle=90:15pt]}] ([xshift=0.5cm] AuxBelowTimesLeftmost.east)--node[above]{\Huge Time}++(AuxBelowTimesRightmost) ;

\end{tikzpicture}
}
}
\end{myframe}
\caption{Schematic review of main information extraction methods discussed in this survey, representing the evolution of algorithms through time. Moving from left to right, the methods are sorted in a  chronological order. This figure shows that recent algorithms are applicable in more general cases (but usually provide less theoretical guarantees).}
\label{fig:summaryTreeOfMethods2}
\end{figure*}

Another concept to note is that despite the connection between IB and VAE, some of VAE issues that researchers have reported do not directly apply to IB. In fact, we think that it is helpful to use the IB interpretation to understand the VAE problems to remedy them. 
For example, one of the improvements over the original VAE, is $\beta$-VAE \cite{HigginsbetaVAELearningBasic2016}. They found that having $\beta>1$ leads to a better performance compared to the original configuration of VAE which is restricted to $\beta=1$.
This phenomena can be studied by using its counterpart results in IB. As mentioned in Section \ref{sec_part:discussionAboutBiasVarianceAndBestBeta}, $\beta$ controls the bias-variance trade-off in case of finite training set. Therefore, one should search for $\beta^*$ which practically does the best in preventing the model from over-fitting. The same argument might be applied to VAE.

Another issue in VAE, which has attracted the attention of many researchers \citep{AlemiFixingBrokenELBO2018,ChenVariationallossyautoencoder2016,ZhaoInfovaeInformationmaximizing2017,ZhengDegenerationVAELight2018} , is that when the family of decoders $p_\theta(x|\tilde{x})$ is too powerful, the loss function \eqref{eq:variational} can be minimized by just using the decoder and completely ignoring the latent variable; i.e. $p_\theta(x|\tilde{x})=p(x)$. In this case, the optimization function \eqref{eq:variational} will be decomposed into two separate terms, where the first term just depends on $\theta$ and the second term just depends on $\phi$. As a result, the second term will be minimized by setting $p_\phi(\tilde{x}|x)=p_0(\tilde{x})$\footnote{Note that usually $p_0$ is a member of the family that is parameterized by $\phi$ (and is actually one of the simplest members). But even if $\nexists \phi: p_\phi(\tilde{x}|x)=p_0(\tilde{x})$, the optimization process does not enforce any relation between $\tilde{x}$ and $x$, and it is enough for our discussion.}. Therefore, $x$ and $\tilde{x}$ will be independent, which is obviously not desired in a feature extraction problem. This problem does not exist in the original IB formulation, in which the focus is on $p_\phi(\tilde{x}|x)$ and
$p(x|\tilde{x})\propto p(\tilde{x}|x)p(x)$ 
is computed without any degrees-of-freedom (no parameter $\theta$ to optimize). It is in contrast with the VAE settings where the discussion starts from $p_\theta(x|\tilde{x})$ and later $p_\phi(\tilde{x}|x)$ is introduced in variational inference. Note that having a strong family of encoders $p_\phi(\tilde{x}|x)$, does not make any problem as long as it is adequately regularized by $\KLdiv{p_\phi(\tilde{x}|x^{(i)})}{p_0(\tilde{x})}$. It should be added that even though IB does not inherently suffer from the "too strong decoder" problem, the current methods which are based on the variational distribution and optimization of both $\theta$ and $\phi$ are not immune to it \cite{AlemiDeepVariationalInformation2017,AchilleEmergenceInvarianceDisentangling2017,AchilleInformationDropoutLearning2018}. This is currently an active research area and we believe the IB viewpoint will help to develop better solutions to it.

In Figure \ref{fig:summaryTreeOfMethods2}, the summary of existing methods and how they evolved trough time, is represented in a hierarchical structure. Note that the solution based on variational techniques \cite{AchilleInformationDropoutLearning2018} bypasses all the limitations that are faced in previous sections; i.e., meaning that it is not limited to a specific family of distributions, does not need the distribution function to be known, and also works for continuous variables. As it is represented in this figure, while the recent methods are capable of solving more general problems, the theoretical guarantees for them are more scarce.%

\section{Beyond Information Bottleneck}
\label{sec:beyondInformationBottleneck}
All the methods discussed till now were using IB which uses the quantity $I(X;T)$ to control the variance of the method (see Section \ref{sec_part:discussionAboutBiasVarianceAndBestBeta}). While this approach is used successfully in many applications, its complete theoretical analysis in the general case is difficult. In this section, a different approach based on mutual information which recently has attracted the attention of researchers is presented. In this new view, instead of looking at $I(X;T)$ as the notion of complexity, one considers $I(S;\mathcal{A}(S))$. Here $S$ is the set of all training samples, and $\mathcal{A}$ is the learning algorithm which uses training points to calculate a hypothesis $h$.

In this approach, not only the mutual information of a single sample $X$ and its representation is considered, but also the mutual information between all of the samples and the whole learned model is studied. 

Following recent information theoretic techniques from \cite{RussoHowmuchdoes2015,RussoControllingbiasadaptive2016,XuInformationtheoreticanalysisgeneralization2017}, authors of paper \cite{BassilyLearnersthatUse2018} used the following notion to prove the interesting inequality 

\begin{equation}
\label{eq:learnersThatUseLittleInformation}
P\left[ \left| \text{err}_{\text{test}} - \text{err}_{\text{train}} \right| > \epsilon \right] < O\left(\frac{I(S;\mathcal{A}(S))}{n\epsilon^2}\right),
\end{equation}
where $\text{err}_{\text{test}}$ and $\text{err}_{\text{train}}$ are the test (true) error and the training (empirical) error of the hypothesis $\mathcal{A}(S)$, respectively, $n$ is the training size, and $\epsilon>0$ is a positive real number.

The intuition behind this inequality is that, the more a learning algorithm uses bits of the training set, %
there is potentially more risk that it will overfit to it. 
The interesting property of this inequality  is that the mutual information between the whole input and output of the algorithm, depends deeply on all the aspects of the learning algorithm. It is in contrast with many other approaches that use the properties of the hypotheses space $\mathcal{H}$ to bound the generalization gap, and usually the effect of final hypothesis chosen by the learning algorithm is blurred away due to the usage of a uniform convergence in proving bounds; like in the Vapnik-Chervonenkis theory \cite{VapnikUniformConvergenceRelative1971}. In paper \cite{AsadiChainingMutualInformation2018}, the chaining method \cite{Dudleysizescompactsubsets1967} was used to further improve the inequality (\ref{eq:learnersThatUseLittleInformation}) to also take into account the capacity of the hypotheses space.

Though the inequality \refeq{eq:learnersThatUseLittleInformation} seems appealing as it directly bounds the generalization error by the simple-looking information theoretic term $I(S;\mathcal{A}(S))$, unfortunately the calculation/estimation of this term is even harder than $I(X;T)$ which was used in IB. This made it quite challenging to apply this technique in real world machine learning problems where the distribution is unknown and the learning algorithms is usually quite complex \cite{VeraRoleInformationBottleneck2018,VeraRoleInformationComplexity2018}. 

To the best knowledge of the authors, the only attempt made to use this technique to analyze the deep learning process is the recent article \cite{ZhangInformationTheoreticViewDeep2018}. In that work, authors argue that as the dataset $S$ goes trough DNN layers $1 \dots m$, the intermediate sequence of datasets $(S_\ell)_{\ell=1}^m$ are formed and $I(S_\ell;W)$ is a decreasing function of $\ell$ (here $W$ is the set of all weights in the DNN). They further argue that this can be used along the inequality (\ref{eq:learnersThatUseLittleInformation}) to show that deeper architectures have less generalization error. A major problem with their analysis is that they used the Markov assumption 
$W$ -- $S$ -- $S_1$ -- $S_2$ ... $S_{m-1}$ -- $S_m$. This assumption does not generally hold in a DNN. Because for calculating the $S_\ell$, a direct usage of $W$ is needed (more precisely the weights up to layer $\ell$ are used). Therefore, it seems that the correct application of this technique in analyzing DNNs requires a more elaborate treatment which is hopped to be released in near future.

\section{Conclusion}
A survey on the interaction of IB and DNNs was given. First, the headlines of the prolong history of using the information theory in ML was presented. The focus was on how the ideas evolved over time. The discussion started from MSS which is practically restricted to distributions from exponential family. Then the IB framework and the Blahut-Arimoto algorithm were discussed which do not work for unknown continuous distributions. After that methods based on variational approximation introduced which are applicable to quite general cases. Finally, another more theoretically appealing usage of information theory was introduced, which used the mutual information between the training set and the learned model to bound the generalization error of a learning algorithm. Despite its theoretical benefits, it was shown that its application in understanding DNNs, is challenging.

During this journey, it was revealed that how some seemingly unrelated areas have hidden relations to the IB. It was also shown that how the mysterious generalization power of SGD (which is the De facto learning method of DNNs) is hypothesized to be caused by the implicit IB compression property which is hidden in SGD. Also, the recent successful unsupervised method VAE was found to be a special case of the IB when solved by employing the variational approximation.

In fact, the profound and seemingly simple tools that the information theory provides bring some traps. As the understanding of these pitfalls are as important, they were also discussed in this survey. It could be seen that how seemingly harmless information theoretic formulas can make impossible situations. Two major  discussed cases were: i) using the mutual information to train continuous deterministic DNNs, which made the problem ill-posed, and ii) using variational approximations without restricting the space of solutions can easily result in meaningless situations. 
The important lesson learned from these revelations was how the ideas from the information theory can give a unified view to different ML concepts. We believe that this view is quite helpful to understand the shortcomings of methods and to remedy them. %

\label{sec:conclusion}

\section*{Acknowledgment}
\AcknowledgmentText

\vskip 0.2in
\IfFileExists{../WholeLibrary_BetterBibTex.bib}{\bibliography{../WholeLibrary_BetterBibTex.bib}}{\bibliography{WholeLibrary_BetterBibTex.bib}}

\begin{thebibliography}{10}

\bibitem{Shannonmathematicaltheorycommunication1948}
C.~E. Shannon, ``A mathematical theory of communication,'' {\em Bell system
  technical journal}, vol.~27, no.~3, pp.~379--423, 1948.

\bibitem{Kullbackinformationsufficiency1951}
S.~Kullback and R.~A. Leibler, ``On information and sufficiency,'' {\em The
  annals of mathematical statistics}, vol.~22, no.~1, pp.~79--86, 1951.

\bibitem{BassilyLearnersthatUse2018}
R.~Bassily, S.~Moran, I.~Nachum, J.~Shafer, and A.~Yehudayoff, ``Learners that
  {{Use Little Information}},'' in {\em Algorithmic {{Learning Theory}}},
  pp.~25--55, 2018.

\bibitem{VeraRoleInformationComplexity2018}
M.~Vera, P.~Piantanida, and L.~R. Vega, ``The {{Role}} of {{Information
  Complexity}} and {{Randomization}} in {{Representation Learning}},'' {\em
  arXiv:1802.05355 [cs, stat]}, Feb. 2018.

\bibitem{NachumDirectSumResult2018}
I.~Nachum, J.~Shafer, and A.~Yehudayoff, ``A {{Direct Sum Result}} for the
  {{Information Complexity}} of {{Learning}},'' {\em arXiv:1804.05474 [cs,
  math, stat]}, Apr. 2018.

\bibitem{SrivastavaDropoutsimpleway2014}
N.~Srivastava, G.~Hinton, A.~Krizhevsky, I.~Sutskever, and R.~Salakhutdinov,
  ``Dropout: {{A}} simple way to prevent neural networks from overfitting,''
  {\em The Journal of Machine Learning Research}, vol.~15, no.~1,
  pp.~1929--1958, 2014.

\bibitem{ZhangUnderstandingdeeplearning2017}
C.~Zhang, S.~Bengio, M.~Hardt, B.~Recht, and O.~Vinyals, ``Understanding deep
  learning requires rethinking generalization,'' {\em International Conference
  on Learning Representations}, 2017.

\bibitem{TishbyDeepLearningInformation2015}
N.~Tishby and N.~Zaslavsky, ``Deep {{Learning}} and the {{Information
  Bottleneck Principle}},'' {\em arXiv preprint arXiv:1503.02406}, 2015.

\bibitem{Tishbyinformationbottleneckmethod1999}
N.~Tishby, F.~Pereira, and W.~Bialek, ``The information bottleneck method,'' in
  {\em Proceedings of the 37-th {{Annual Allerton Conference}} on
  {{Communication}}, {{Control}} and {{Computing}}}, pp.~368--377, 1999.

\bibitem{Shwartz-ZivOpeningBlackBox2017}
R.~{Shwartz-Ziv} and N.~Tishby, ``Opening the {{Black Box}} of {{Deep Neural
  Networks}} via {{Information}},'' {\em arXiv:1703.00810 [cs]}, Mar. 2017.

\bibitem{KhadiviFlowinformationfeedforward2016}
P.~Khadivi, R.~Tandon, and N.~Ramakrishnan, ``Flow of information in
  feed-forward deep neural networks,'' {\em arXiv preprint arXiv:1603.06220},
  2016.

\bibitem{AchilleEmergenceInvarianceDisentangling2017}
A.~Achille and S.~Soatto, ``On the {{Emergence}} of {{Invariance}} and
  {{Disentangling}} in {{Deep Representations}},'' {\em arXiv:1706.01350 [cs,
  stat]}, June 2017.

\bibitem{SaxeInformationBottleneckTheory2018}
A.~M. Saxe, Y.~Bansal, J.~Dapello, M.~Advani, A.~Kolchinsky, B.~D. Tracey, and
  D.~D. Cox, ``On the {{Information Bottleneck Theory}} of {{Deep Learning}},''
  {\em International Conference on Learning Representations}, Feb. 2018.

\bibitem{AlemiDeepVariationalInformation2017}
A.~A. Alemi, I.~Fischer, J.~V. Dillon, and K.~Murphy, ``Deep {{Variational
  Information Bottleneck}},'' {\em International Conference on Learning
  Representations}, 2017.

\bibitem{KolchinskyNonlinearInformationBottleneck2017}
A.~Kolchinsky, B.~D. Tracey, and D.~H. Wolpert, ``Nonlinear {{Information
  Bottleneck}},'' {\em arXiv:1705.02436 [cs, math, stat]}, May 2017.

\bibitem{AchilleInformationDropoutLearning2018}
A.~Achille and S.~Soatto, ``Information {{Dropout}}: {{Learning Optimal
  Representations Through Noisy Computation}},'' {\em IEEE Transactions on
  Pattern Analysis and Machine Intelligence}, pp.~1--1, 2018.

\bibitem{KolmogorovShannontheoryinformation1956}
A.~Kolmogorov, ``On the {{Shannon}} theory of information transmission in the
  case of continuous signals,'' {\em IRE Transactions on Information Theory},
  vol.~2, no.~4, pp.~102--108, 1956.

\bibitem{CoverKolmogorovContributionsInformation1989}
T.~M. Cover, P.~Gacs, and R.~M. Gray, ``Kolmogorov's {{Contributions}} to
  {{Information Theory}} and {{Algorithmic Complexity}},'' {\em The annals of
  probability}, no.~3, pp.~840--865, 1989.

\bibitem{CoverElementsinformationtheory2012}
T.~M. Cover and J.~A. Thomas, {\em Elements of Information Theory}.
\newblock {John Wiley \& Sons}, 2012.

\bibitem{RAFishermathematicalfoundationstheoretical1922}
M.~A. RA~Fisher, ``On the mathematical foundations of theoretical statistics,''
  {\em Phil. Trans. R. Soc. Lond. A}, vol.~222, no.~594-604, pp.~309--368,
  1922.

\bibitem{ShamirLearninggeneralizationinformation2010}
O.~Shamir, S.~Sabato, and N.~Tishby, ``Learning and generalization with the
  information bottleneck,'' {\em Theoretical Computer Science}, vol.~411,
  pp.~2696--2711, June 2010.

\bibitem{LehmannCompletenessSimilarRegions1948}
E.~L. Lehmann and H.~Scheff\'e, ``Completeness, {{Similar Regions}}, and
  {{Unbiased Estimation}},'' in {\em Bulletin of the {{American Mathematical
  Society}}}, vol.~54, pp.~1080--1080, {Charles St, Providence}, 1948.

\bibitem{Koopmandistributionsadmittingsufficient1936}
B.~O. Koopman, ``On distributions admitting a sufficient statistic,'' {\em
  Transactions of the American Mathematical society}, vol.~39, no.~3,
  pp.~399--409, 1936.

\bibitem{FriedmanMultivariateinformationbottleneck2001}
N.~Friedman, O.~Mosenzon, N.~Slonim, and N.~Tishby, ``Multivariate information
  bottleneck,'' in {\em Proceedings of the {{Seventeenth}} Conference on
  {{Uncertainty}} in Artificial Intelligence}, pp.~152--161, {Morgan Kaufmann
  Publishers Inc.}, 2001.

\bibitem{Gilad-Bachrachinformationtheoretictradeoff2003}
R.~{Gilad-Bachrach}, A.~Navot, and N.~Tishby, ``An information theoretic
  tradeoff between complexity and accuracy,'' in {\em Learning {{Theory}} and
  {{Kernel Machines}}}, pp.~595--609, {Springer}, 2003.

\bibitem{BlahutComputationchannelcapacity1972}
R.~Blahut, ``Computation of channel capacity and rate-distortion functions,''
  {\em IEEE transactions on Information Theory}, vol.~18, no.~4, pp.~460--473,
  1972.

\bibitem{Arimotoalgorithmcomputingcapacity1972}
S.~Arimoto, ``An algorithm for computing the capacity of arbitrary discrete
  memoryless channels,'' {\em IEEE Transactions on Information Theory},
  vol.~18, no.~1, pp.~14--20, 1972.

\bibitem{ChechikInformationbottleneckGaussian2005}
G.~Chechik, A.~Globerson, N.~Tishby, and Y.~Weiss, ``Information bottleneck for
  {{Gaussian}} variables,'' {\em Journal of machine learning research}, vol.~6,
  no.~Jan, pp.~165--188, 2005.

\bibitem{KrizhevskyImagenetclassificationdeep2012}
A.~Krizhevsky, I.~Sutskever, and G.~E. Hinton, ``Imagenet classification with
  deep convolutional neural networks,'' in {\em Advances in Neural Information
  Processing Systems}, pp.~1097--1105, 2012.

\bibitem{Keskarlargebatchtrainingdeep2016}
N.~S. Keskar, D.~Mudigere, J.~Nocedal, M.~Smelyanskiy, and P.~T.~P. Tang, ``On
  large-batch training for deep learning: {{Generalization}} gap and sharp
  minima,'' {\em arXiv preprint arXiv:1609.04836}, 2016.

\bibitem{HochreiterFlatminima1997}
S.~Hochreiter and J.~Schmidhuber, ``Flat minima,'' {\em Neural Computation},
  vol.~9, no.~1, pp.~1--42, 1997.

\bibitem{ChaudhariEntropysgdBiasinggradient2016}
P.~Chaudhari, A.~Choromanska, S.~Soatto, and Y.~LeCun, ``Entropy-sgd:
  {{Biasing}} gradient descent into wide valleys,'' {\em arXiv preprint
  arXiv:1611.01838}, 2016.

\bibitem{HardtTrainfastergeneralize2015}
M.~Hardt, B.~Recht, and Y.~Singer, ``Train faster, generalize better:
  {{Stability}} of stochastic gradient descent,'' {\em arXiv preprint
  arXiv:1509.01240}, 2015.

\bibitem{PoggioWhyWhenCan2016}
T.~Poggio, H.~Mhaskar, L.~Rosasco, B.~Miranda, and Q.~Liao, ``Why and {{When
  Can Deep}}\textendash{}but {{Not Shallow}}\textendash{{Networks Avoid}} the
  {{Curse}} of {{Dimensionality}},'' {\em arXiv preprint arXiv:1611.00740},
  2016.

\bibitem{DinhSharpminimacan2017}
L.~Dinh, R.~Pascanu, S.~Bengio, and Y.~Bengio, ``Sharp minima can generalize
  for deep nets,'' {\em arXiv preprint arXiv:1703.04933}, 2017.

\bibitem{VidalMathematicsDeepLearning2017}
R.~Vidal, J.~Bruna, R.~Giryes, and S.~Soatto, ``Mathematics of {{Deep
  Learning}},'' {\em arXiv:1712.04741 [cs]}, Dec. 2017.

\bibitem{AmjadHowNotTrain2018}
R.~A. Amjad and B.~C. Geiger, ``How ({{Not}}) {{To Train Your Neural Network
  Using}} the {{Information Bottleneck Principle}},'' {\em arXiv:1802.09766
  [cs, math]}, Feb. 2018.

\bibitem{ChaudhariStochasticgradientdescent2017}
P.~Chaudhari and S.~Soatto, ``Stochastic gradient descent performs variational
  inference, converges to limit cycles for deep networks,'' {\em
  arXiv:1710.11029 [cond-mat, stat]}, Oct. 2017.

\bibitem{KingmaAutoencodingvariationalbayes2013}
D.~P. Kingma and M.~Welling, ``Auto-encoding variational bayes,'' {\em arXiv
  preprint arXiv:1312.6114}, 2013.

\bibitem{BengioGeneralizeddenoisingautoencoders2013}
Y.~Bengio, L.~Yao, G.~Alain, and P.~Vincent, ``Generalized denoising
  auto-encoders as generative models,'' in {\em Advances in {{Neural
  Information Processing Systems}}}, pp.~899--907, 2013.

\bibitem{ImDenoisingCriterionVariational2017}
D.~J. Im, S.~Ahn, R.~Memisevic, and Y.~Bengio, ``Denoising {{Criterion}} for
  {{Variational Auto}}-{{Encoding Framework}}.,'' in {\em {{AAAI}}},
  pp.~2059--2065, 2017.

\bibitem{AlemiFixingBrokenELBO2018}
A.~A. Alemi, B.~Poole, I.~Fischer, J.~V. Dillon, R.~A. Saurous, and K.~Murphy,
  ``Fixing a {{Broken ELBO}},'' {\em arXiv:1711.00464 [cs, stat]}, Feb. 2018.

\bibitem{ChenVariationallossyautoencoder2016}
X.~Chen, D.~P. Kingma, T.~Salimans, Y.~Duan, P.~Dhariwal, J.~Schulman,
  I.~Sutskever, and P.~Abbeel, ``Variational lossy autoencoder,'' {\em arXiv
  preprint arXiv:1611.02731}, 2016.

\bibitem{HuangFlowRenyiinformation2016}
C.-W. Huang and S.~S.~S. Narayanan, ``Flow of {{Renyi}} information in deep
  neural networks,'' in {\em Machine {{Learning}} for {{Signal Processing}}
  ({{MLSP}}), 2016 {{IEEE}} 26th {{International Workshop}} On}, pp.~1--6,
  {IEEE}, 2016.

\bibitem{HigginsbetaVAELearningBasic2016}
I.~Higgins, L.~Matthey, A.~Pal, C.~Burgess, X.~Glorot, M.~Botvinick,
  S.~Mohamed, and A.~Lerchner, ``Beta-{{VAE}}: {{Learning Basic Visual
  Concepts}} with a {{Constrained Variational Framework}},'' {\em International
  Conference on Learning Representations}, Nov. 2016.

\bibitem{ZhaoInfovaeInformationmaximizing2017}
S.~Zhao, J.~Song, and S.~Ermon, ``Infovae: {{Information}} maximizing
  variational autoencoders,'' {\em arXiv preprint arXiv:1706.02262}, 2017.

\bibitem{ZhengDegenerationVAELight2018}
H.~Zheng, J.~Yao, Y.~Zhang, and I.~W. Tsang, ``Degeneration in {{VAE}}: In the
  {{Light}} of {{Fisher Information Loss}},'' {\em arXiv:1802.06677 [cs,
  stat]}, Feb. 2018.

\bibitem{RussoHowmuchdoes2015}
D.~Russo and J.~Zou, ``How much does your data exploration overfit?
  {{Controlling}} bias via information usage,'' {\em arXiv preprint
  arXiv:1511.05219}, 2015.

\bibitem{RussoControllingbiasadaptive2016}
D.~Russo and J.~Zou, ``Controlling bias in adaptive data analysis using
  information theory,'' in {\em Artificial {{Intelligence}} and
  {{Statistics}}}, pp.~1232--1240, 2016.

\bibitem{XuInformationtheoreticanalysisgeneralization2017}
A.~Xu and M.~Raginsky, ``Information-theoretic analysis of generalization
  capability of learning algorithms,'' in {\em Advances in {{Neural Information
  Processing Systems}}}, pp.~2521--2530, 2017.

\bibitem{VapnikUniformConvergenceRelative1971}
V.~N. Vapnik and A.~Y. Chervonenkis, ``On the {{Uniform Convergence}} of
  {{Relative Frequencies}} of {{Events}} to {{Their Probabilities}},'' {\em
  Theory of Probability and its Applications}, vol.~16, no.~2, p.~264, 1971.

\bibitem{AsadiChainingMutualInformation2018}
A.~R. Asadi, E.~Abbe, and S.~Verd\'u, ``Chaining {{Mutual Information}} and
  {{Tightening Generalization Bounds}},'' {\em arXiv preprint
  arXiv:1806.03803}, 2018.

\bibitem{Dudleysizescompactsubsets1967}
R.~M. Dudley, ``The sizes of compact subsets of {{Hilbert}} space and
  continuity of {{Gaussian}} processes,'' {\em Journal of Functional Analysis},
  vol.~1, no.~3, pp.~290--330, 1967.

\bibitem{VeraRoleInformationBottleneck2018}
M.~Vera, P.~Piantanida, and L.~R. Vega, ``The {{Role}} of the {{Information
  Bottleneck}} in {{Representation Learning}},'' in {\em 2018 {{IEEE
  International Symposium}} on {{Information Theory}} ({{ISIT}})}, (Vail, CO),
  pp.~1580--1584, {IEEE}, June 2018.

\bibitem{ZhangInformationTheoreticViewDeep2018}
J.~Zhang, T.~Liu, and D.~Tao, ``An {{Information}}-{{Theoretic View}} for
  {{Deep Learning}},'' {\em arXiv preprint arXiv:1804.09060}, 2018.

\end{thebibliography}

\authorsBio

\end{document}